\documentclass[11pt]{article}

\usepackage[preprint]{acl}

\usepackage{times}
\usepackage{latexsym}
\usepackage{natbib}
\setcitestyle{numbers,square}

\usepackage[T1]{fontenc}

\usepackage[utf8]{inputenc}

\usepackage{microtype}

\usepackage{inconsolata}

\usepackage{graphicx}
\usepackage{tcolorbox}

\usepackage{hyperref}       
\usepackage{url}            
\usepackage{booktabs}       
\usepackage{amsfonts}       
\usepackage{nicefrac}       
\usepackage{microtype}      
\usepackage{xcolor}         
\usepackage{graphicx} 
\usepackage{colortbl}
\usepackage{subfigure}
\usepackage{tcolorbox}
\usepackage{wrapfig}
\usepackage{makecell}

\usepackage{breqn}

\usepackage{amssymb} 
\usepackage{amsmath}
\usepackage{caption}

\usepackage{algorithm} 
\usepackage{algorithmic}

\newtheorem{finding}{Finding}

\title{LazyEviction: Lagged KV Eviction with Attention Pattern Observation for Efficient Long Reasoning}

\author{
 \textbf{Haoyue Zhang\textsuperscript{1}},
 \textbf{Hualei Zhang\textsuperscript{1}},
 \textbf{Xiaosong Ma\textsuperscript{2}},
 \textbf{Jie Zhang\textsuperscript{1}},
 \textbf{Song Guo\textsuperscript{1}},
\\
\\
 \textsuperscript{1}HKUST,
 \textsuperscript{2}HK PolyU
\\
 \small{
   \textbf{Correspondence:Jie Zhang,} \href{mailto:email@domain}{csejzhang@ust.hk}
 }
}

\begin{document}
\maketitle
\begin{abstract}
Large Language Models (LLMs) exhibit enhanced capabilities by Chain-of-Thought reasoning. However, the extended reasoning sequences introduce significant GPU memory overhead due to increased key-value (KV) cache. Existing KV cache compression methods mitigate memory bottlenecks but struggle in long reasoning tasks. In this paper, we analyze attention patterns in reasoning tasks and reveal a \textbf{Token Importance Recurrence} phenomenon: a large proportion of tokens regain high attention after multiple decoding steps, which is failed to capture by existing works and may lead to unpredictable eviction on such periodically critical tokens.
%
To address this, we propose \textbf{LazyEviction}, an observation window-based lagged eviction framework retaining latent recurring tokens by prioritized eviction based on tokens' recurrence patterns. 
Extensive experiments demonstrate that LazyEviction reduces KV cache by 50\%\textasciitilde70\% while maintaining comparable accuracy, outperforming existing KV cache compression baselines. Our implementation code can be found at \url{https://github.com/Halo-949/LazyEviction}.

\end{abstract}

\section{Introduction}
Large Language Models (LLMs) have emerged advanced capabilities in complex reasoning tasks by enabling Chain-of-Thought (CoT) to elicit step-by-step inference~\citep{wei2022chain,kojima2022large}. Recent advancements, such as OpenAI’s o1~\citep{jaech2024openai} and DeepSeekR1~\citep{guo2025deepseek}, further demonstrate that scaling up CoT lengths from hundreds to thousands of reasoning steps could continuously improve LLM reasoning. 
However, the increased reasoning sequences introduce substantial memory and computational overhead. Due to the autoregressive nature of LLM decoding, longer CoT outputs lead to proportional increases in GPU memory for caching key-value (KV) states. These issues become particularly pronounced when reasoning sequences extend into thousands of reasoning tokens, resulting in significant GPU memory costs. For example, in mathematics~\citep{cobbe2021training, hendrycks2021measuring} and programming tasks~\citep{chen2021evaluating}, the reasoning sequences can grow up to 16k tokens, resulting in more than 100GB KV cache with a batch size of 32, which exceeds the memory capacity of even high-end GPUs.

\begin{figure*}[tb]
\centering{\includegraphics[scale=0.42]{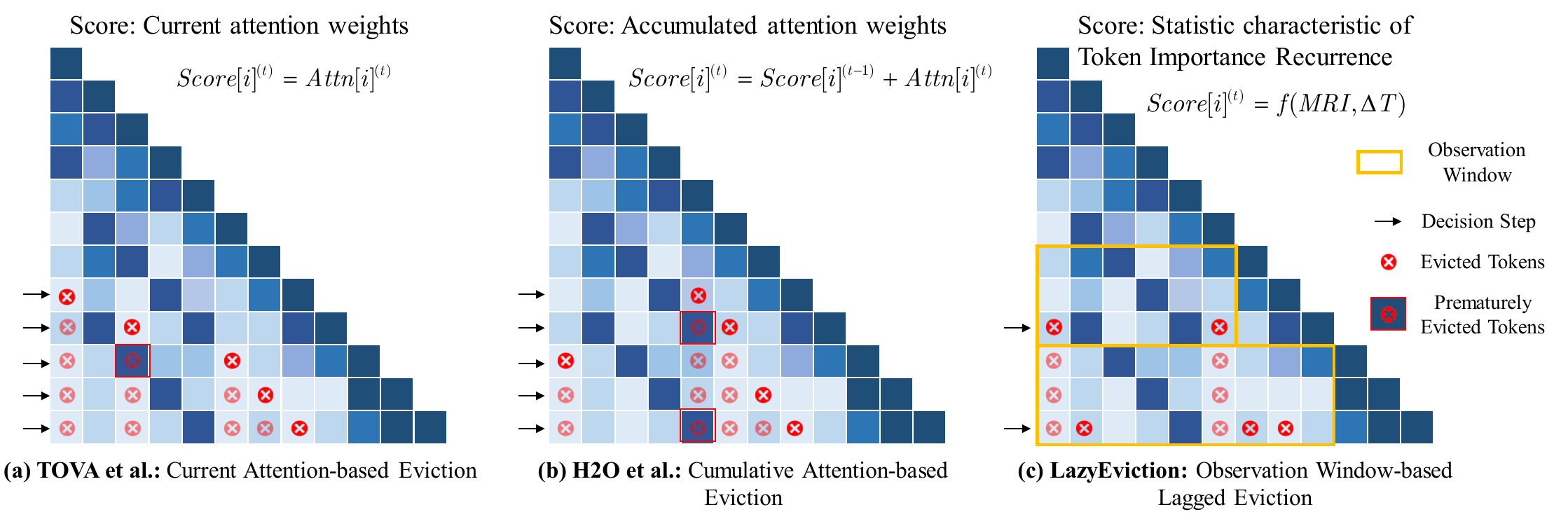}}
\caption{Comparison of different KV Eviction methods. The dark squares represent that the token has a higher attention score. (a) Current Attention-based Eviction executes stepwise evictions using immediate attention scores. (b) Cumulative Attention-based Eviction integrates historical attention for eviction decisions. Both (a) and (b) fail to preserve recurring tokens during their low-attention intervals. (c) LazyEviction performs lagged KV evictions based on the observation window to detect latent recurring tokens and prevent prematurely discarding them.}
\label{KV}
\vspace{-5mm}
\end{figure*}

KV cache compression has emerged as a promising approach to alleviate the memory bottleneck caused by increasing KV cache sizes. Unlike traditional long-context input tasks that focus on the prefilling phase~\citep{li2024snapkv,cai2024pyramidkv,feng2024ada,fu2024not}, long-reasoning tasks perform long-context generation that requires compression in multiple decoding steps. The critical challenge lies in compressing generated KV caches while minimizing performance degradation. Existing approaches mitigate this through selective eviction strategies: early work~\citep{xiao2023efficient} preserves subsets of recent and initial KV pairs, while others~\citep{oren2024transformers,chen2024nacl,he2025treekv,zhang2023h2o,adnan2024keyformer} leverage attention scores to prioritize critical tokens and evict others, which can be further divided into two types: 
(a) Current Attention-based Eviction \citep{oren2024transformers, chen2024nacl, he2025treekv} and (b) Cumulative Attention based Eviction \citep{zhang2023h2o, adnan2024keyformer, ghadia2025dialogue, hu2025efficient}. However, as shown in Fig.~\ref{KV}, these works are greedy eviction strategies at each decoding step, which discard tokens deemed temporarily unimportant while overlooking their potential future importance. 


We empirically analyze attention patterns during multi-step reasoning processes. Our investigation reveals a frequent \textbf{Token Importance Recurrence (TIR)} phenomenon: a large proportion of tokens in reasoning tasks receive renewed attention after multiple decoding steps. In this paper, we refer to such tokens as \textbf{recurring tokens}, where the attention score of recurring tokens may be low within an interval while suddenly increasing in the future. Empirical analysis demonstrates that these tokens typically involve conditional or summary information within reasoning chains, aligning with the capabilities in verification, backtracking, and summarization in reasoning models~\citep{gandhi2025cognitive}. The recurring tokens often contain information vital for later reasoning stages, and premature eviction of recurring tokens leads to catastrophic performance degradation. Crucially, existing KV compression methods usually fail to identify such tokens due to inherent limitations, thus risking prematurely discarding crucial tokens during reasoning steps, ultimately sacrificing task accuracy.


To address this challenge, we propose LazyEviction, a novel \textbf{Observation Window-based Lagged Eviction Mechanism}. Instead of per-step greedy decisions, we perform lagged KV evictions at every $W$ decoding steps to detect latent recurring tokens and prevent prematurely discarding them. There are two key components: (1) \textbf{Recurrence Interval Tracking:} we record each token's Maximum Recurrence Interval (MRI) (i.e., the longest observed period between its attention spikes) to capture the temporal variation of token importance. (2) \textbf{MRI-Centric Eviction Policy:} we prioritize evicting tokens whose $\Delta T$ exceeds their MRI, where $\Delta T$ is the time elapsed since the token's last important recurrence. Tokens with $\Delta T<MRI$ are retained as they still have the potential to become important in the future. LazyEviction shifts from step-wised greedy eviction to window-wised predictive retention: by continuously tracking MRI and executing lagged eviction during observation windows, we can effectively detect latent recurring tokens and prevent prematurely discarding them. To sum up, our key contributions are:

\begin{itemize}
    \item We observe token importance recurrence in the reasoning task, where specific tokens exhibit recurrent attention patterns. These recurring tokens often represent foundational formulations or intermediate conclusions that are crucial for maintaining knowledge continuity in multi-step reasoning.
    \item We propose LazyEviction, an observation window-based lagged KV eviction scheme integrating recurrence interval tracking. Our framework enables prediction of future token importance, thereby preventing premature eviction of recurring tokens.
    \item We have implemented LazyEviction on the latest large reasoning models ranging from 4B to 32B. It has been verified on multiple reasoning datasets that our method can maintain comparable performance while reducing 50\%\textasciitilde70\% KV budget, surpassing the SOTA KV cache compression methods. 
\end{itemize}

\section{Related Works}

Various approaches have been developed to improve LLM's efficiency in handling long reasoning tasks, which can be classified into three main categories: KV cache compression, reasoning path compression, and system-level optimizations.


\noindent\textbf{KV cache compression.} Considering the inherent sparsity of attention mechanisms, these works focus on retaining critical tokens' KV pairs while evicting non-essential ones within constrained memory budgets. We summarize the existing works as follows:

(1) \textit{Static-Position retention}: StreamingLLM ~\citep{xiao2023efficient} employs static retention of initial and recent tokens under fixed budgets. While sustaining text coherence, their rigid retention policies cannot detect latent recurring tokens.

(2) \textit{Current Attention-based Eviction}: TOVA ~\citep{oren2024transformers}, NACL~\citep{chen2024nacl}, and TreeKV~\citep{he2025treekv} execute stepwise evictions using immediate attention scores. This myopic strategy fails to preserve recurring tokens during low-attention intervals.

(3) \textit{Cumulative Attention-based Eviction}: Scissorhands ~\citep{liu2023scissorhands}, H2O~\citep{zhang2023h2o}, Keyformer~\citep{adnan2024keyformer}, and MorphKV~\citep{ghadia2025dialogue} integrate tokens' historical attention values for eviction decisions. RaaS~\citep{hu2025efficient} extends this through dynamic updated timestamp. These approaches still mistakenly regard latent recurring tokens as unimportant due to continuous low historical scores.

Recent works R-KV~\citep{cai2025r} and RPC~\citep{song2025reasoning} utilize token similarity to identify and evict redundant tokens in long reasoning tasks. Instead, LazyEviction directly focuses on the attention patterns in reasoning path. Our goal is to use the compressed KV cache to closely approximate the full KV attention map, ensuring the theoretical interpretability of performance.




\noindent \textbf{Long Reasoning Compression.} 
Recent works have focused on compressing the output of reasoning models. For instance, TALE~\citep{han2024token} and SoT~\citep{aytes2025sketch} adopt prompt engineering to control the length of CoT outputs. TokenSkip~\citep{xia2025tokenskip} achieves output compression through fine-tuning on condensed CoT data. InftyThink~\citep{yan2025inftythink} and LightThinker~\citep{zhang2025lightthinker} equip models with summarization capabilities to reduce tokens during the reasoning step. Distinctively, LazyEviction maintains the original output length of LLMs while enhancing memory and computational efficiency through optimized KV cache usage, demonstrating compatibility with these compression approaches.

\noindent \textbf{System-level Optimizations.} 
By optimizing GPU memory usage and batch processing, vLLM~\citep{kwon2023efficient} and FlashDecoding~\citep{hong2023flashdecoding} imporve both time and memory efficiency in decoding. Additionally, QUEST~\citep{tang2024quest} and OmniKV~\citep{hao2025omnikv} store KV caches in CPU and dynamically schedule them during the decoding phase. Our method is based on an eviction strategy that prioritizes the importance of tokens in the future, and it is parallel and compatible with the above methods.



\section{Observations and Motivations}
\label{observation}

\begin{figure*}[t]
\centering{\includegraphics[scale=0.67]{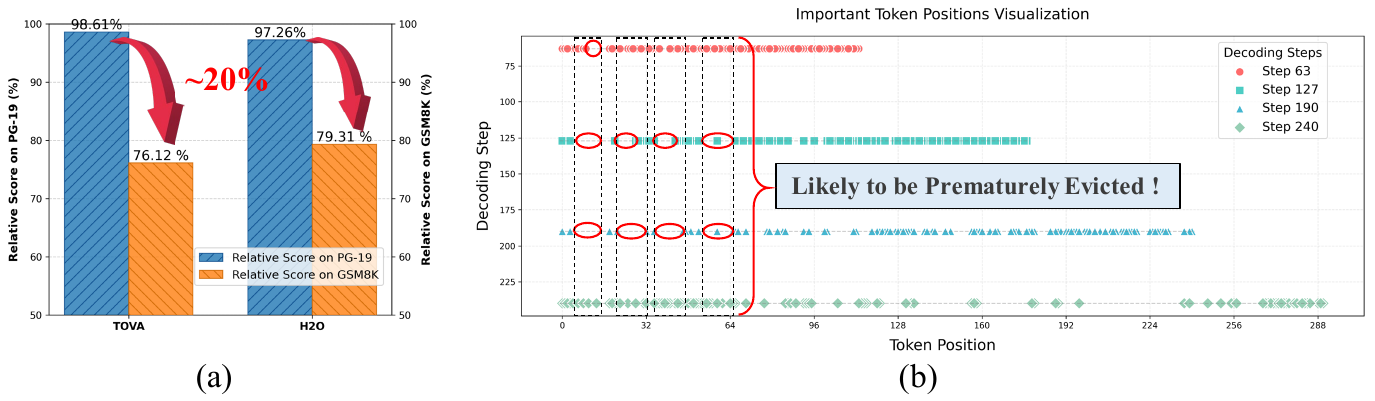}}
\caption{(a) shows the performance degradation for SOTA Methods on reasoning tasks. With the same KV cache compression ratio (e.g., 50\%), compared with traditional language modeling tasks dataset PG-19~\citep{rae2019compressive}, the performance of both H2O and TOVA has decreased by \textasciitilde20\% on GSM8K dataset. (b) is the visualization of the importance variation by selecting Top-50\% important tokens. Tokens at the same position show different importance at different decoding steps.}
\label{bias}
\end{figure*}

\begin{figure*}[t]
\centering{\includegraphics[scale=0.47]{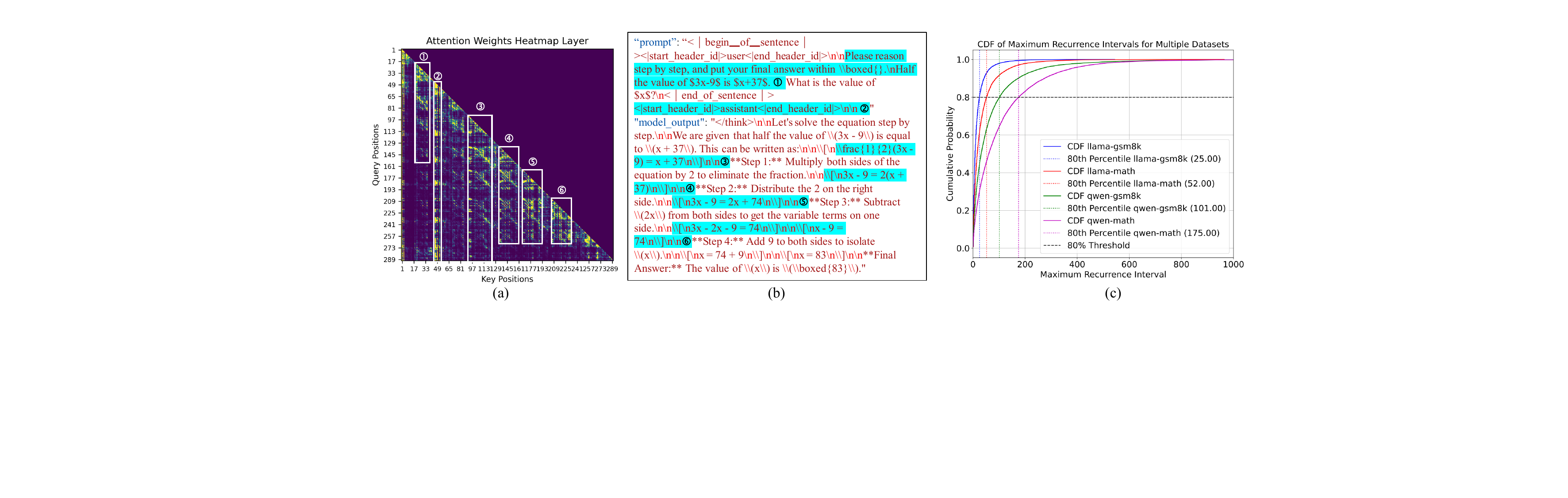}}
\caption{Visualization of TIR. We observe attention maps across different heads of DeepSeek-R1-Distill-Llama-8B. We find most tokens(>95\%) show TIR pattern. (a) and (b) show an attention map with recurring tokens and their corresponding positions. (c) statistically analyzed the MRI distribution in different models among different tasks.}
\label{attn}
\end{figure*}




The goal of KV cache eviction is to identify and evict KV pairs that are non-critical for future decoding steps. To preserve the correctness of subsequent decoding results, it is critical to retain tokens with potential future importance. In this section, we first analyze the previous works and observe attention patterns in reasoning tasks. 

\begin{tcolorbox}
\begin{finding}
Exiting works fail to reserve the potentially important tokens during decoding steps, thus exhibiting degraded performance in long reasoning tasks.
\end{finding}
\end{tcolorbox}

We compare the performance of H2O~\citep{zhang2023h2o} and TOVA~\citep{oren2024transformers} on reasoning tasks versus general long-context generation tasks in Fig.~\ref{bias}(a), H2O and TOVA report competitive performance on standard language modeling dataset, but exhibit significant performance degradation on GSM8K under the same KV cache compression ratio. To diagnose this failure, we track the positions of top-50\% important tokens across decoding steps in Fig.~\ref{bias}(b) and observe that tokens critical in later steps are often missing in earlier steps. Due to their low importance scores in intermediate steps, these tokens are prone to premature eviction in prior works.

\begin{tcolorbox}
\begin{finding}
	Most tokens (>95\%) exhibit recurrent attention patterns in multi-step reasoning tasks. Since they contain critical information for the reasoning, prematurely evicting may cause knowledge incontinuity.
\end{finding}
\end{tcolorbox}

We observe a prevalent phenomenon in reasoning tasks, where tokens initially exhibit strong attention weights, receive low attention weights during later decoding steps, and subsequently regain attention after a certain decoding step interval, as illustrated in Fig.~\ref{attn}(a), \textcircled{\small{1}}–\textcircled{\small{6}}. We define this phenomenon as Token Importance Recurrence (TIR). 

We select a sample from the MATH500 dataset for visualization. We annotate the recurring tokens \textcircled{\small{1}}–\textcircled{\small{6}} from Fig.~\ref{attn}(a) in both the input prompt and output CoT shown in Fig.~\ref{attn}(b). These recurring tokens contain critical information required for reasoning steps, such as the initial problem conditions (tokens \textcircled{\small{1}}) in the prompt, which are repeatedly referenced in subsequent reasoning steps. Additionally, intermediate results (tokens \textcircled{\small{3}}–\textcircled{\small{6}}) generated during reasoning are reactivated in later steps.

To further validate the generality of this phenomenon, we measured the Maximum Recurrence Interval (MRI) — the longest period between consecutive re-activations of a token — across multiple reasoning models on the GSM8K and MATH500 datasets. Results show that over 95\% of tokens exhibit importance recurrence (MRI > 1), confirming its ubiquity in reasoning tasks (Fig.~\ref{attn}(c)).

\begin{tcolorbox}
\begin{finding}
    Most recurring tokens' MRI is far smaller than the output length and can be detected via an observation window.
\end{finding}
\end{tcolorbox}

\begin{figure*}[htbp]
\centering{\includegraphics[scale=0.40]{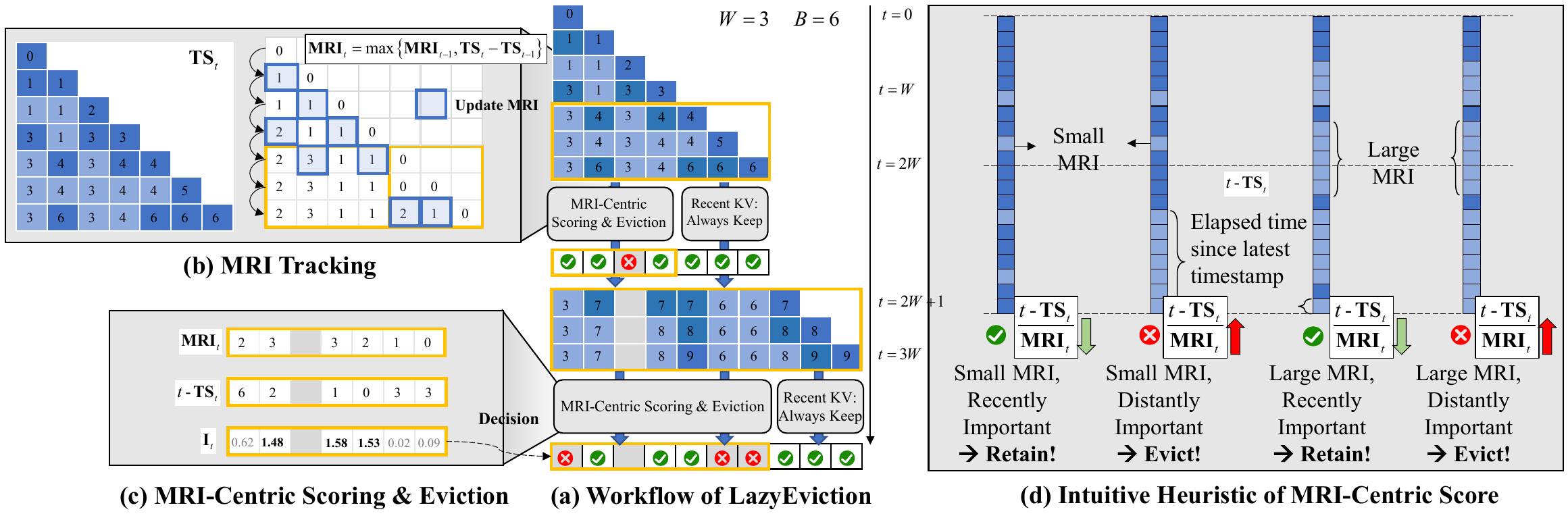}}
\caption{Overview of our proposed LazyEviction Framework, where the dark squares represent that the token has a relatively higher attention score. (a) LazyEviction performs eviction decisions at intervals of $W$ steps. The workflow contains two key operations: (b) Dynamic MRI Tracking according to updated important timestamps, and (c) MRI-Centric Scoring during decision phases, where tokens predicted to be critical for future steps are retained. (d) The MRI-Centric Score fundamentally predicts future token importance by analyzing historical patterns of importance variation (i.e., MRI and the time elapsed since their latest timestamp).}
\label{lazy}
\end{figure*}

As shown in Fig.~\ref{attn}(c), we statistically analyzed the MRI distribution in different models among different tasks. As the output length of the model increases, the MRI gradually becomes larger, but most tokens' MRI remian relatively small. For example, the output length of Qwen model on the MATH500 dataset can reach 8k, but we have statistically found that 80\% tokens' MRI are less than 175. 
Therefore, as long as the window size exceeds the MRI of most tokens, the majority of recurring tokens will be identified. This finding motivates our design of the observation window-based lagged eviction mechanism.




\section{LazyEviction}
\label{sec_lazy}


In this section, we introduce our LazyEviction, a novel framework of \textbf{Observation Window-based Lagged Eviction}.
%
We first model LazyEviction as a multi-step dynamic optimization problem in Appendix~\ref{Appen_formulation}. Unlike existing works that perform KV eviction decisions at every decoding step, our LazyEviction adopts a window size $W$ as the interval for KV eviction decisions. As illustrated in Fig.~\ref{lazy}(a), during decision-making, LazyEviction always retains the most recent $W$ KVs to preserve local coherence. When the number of KVs exceeds the predefined budget $B (B\gg W)$, LazyEviction executes KV eviction. The overall procedure of our LazyEviction is summarized in Appendix.~\ref{algo}.

To select $B-W$ KV pairs from past caches to retain, LazyEviction tracks importance variations (i.e., MRI) of each token at every decoding step (\textbf{Recurrence Interval Tracking}). For eviction decisions, importance scores, which are defined to indicate tokens' potential future importance, are calculated based on MRI. Consequently, LazyEviction selectively retains only the tokens in the KV cache most likely to be critical in the future (\textbf{MRI-Centric Eviction}). 

\paragraph{Recurrence Interval Tracking.}

To identify importance variations for recurring tokens, we employ timestamps to track each token's time of being important. Consistent with RaaS~\citep{hu2025efficient}, whenever a token receives an attention score exceeding a threshold $\alpha$, its latest timestamp is updated to the value of current decoding step $t$. Thus, during the whole decoding phase, LazyEviction dynamically maintains a timestamp vector $\mathbf{TS}_t$ for all retained tokens, which records only the most recent activation time of tokens (illustrated in the leftmost figure in Fig.~\ref{lazy}(b)).

However, only relying on $\mathbf{TS}_t$ fails to capture the temporal patterns of Importance Recurrence. To address this, we introduce MRI to record the longest interval between two consecutive activations in history. For newly generated tokens, MRI is initialized to 0. The MRI value is updated as:
%
%
\begin{equation}
\mathbf{MRI}_t = \max\{\mathbf{MRI}_{t-1}, \mathbf{TS}_t - \mathbf{TS}_{t-1}\}
\label{eq_MRI}.
\end{equation}

\paragraph{MRI-Centric Eviction.}

When the size of cached KV pairs $\left| {\mathcal{S}_t} \right|$ exceeds $B$,  KV eviction are triggered at periodic intervals $t = kW$ ($k \in \mathbf{N}^+$). Leveraging the tracked historical importance dynamics via $\mathbf{MRI}_{t-1}$ and $\mathbf{TS}_t$. We aim to predict each token’s future importance using an MRI-Centric Importance Score. As illustrated in Fig~\ref{lazy}(d), the importance score should be designed under the following two intuitive heuristics:

\begin{enumerate}
    \item \textbf{H1-score:} Given a token $i$, if the time elapsed since its last activation ($t - T{S_t}[i]$) is greater than its $MR{I_t}[i]$, the possibility of token $i$ to be important in the future will decrease. The larger ${\frac{{t - T{S_t}[i]}}{{MR{I_t}[i]}}}$, the less likely token $i$ is important. Thus, we define $\mathbf{H}^{(1)}_t[i] = 2\sigma(-{\frac{{t - T{S_t}[i]}}{{MR{I_t}[i]}}}) $, where $\sigma(x) = 1/(1+\exp(-x))$ is sigmoid function.
    \item \textbf{H2-score:} Tokens with smaller MRI values have a higher possibility to be important in the future, as frequently recurring tokens are deemed more critical. Thus, we define $\mathbf{H}^{(2)}_t[i] = 2\sigma(-{{\frac{1}{{MR{I_t}[i]}-1}}})$. Notably, if $MR{I_t}[i] = 0$, the token has never been activated since its generation (MRI is initialized as 0), resulting in $\mathbf{H}^{(2)}_t[i] = 0, \rm{if}MR{I_t}[i] = 0$.
\end{enumerate}

\textbf{H1-Score} reflects the likelihood of the token regaining importance within the next observation window, while \textbf{H2-Score} prioritizes tokens with smaller MRI values. We discuss the importance of these two score functions in Sec~\ref{score}, and explain the rationale behind the formulation of each function in Appendix~\ref{Appen_formulation}.
Thus, the MRI-Centric Importance Score can be formalized as:
%
%
\begin{equation}
\mathbf{I}_t[i] = \left\{ {\begin{array}{*{20}{c}}
{\mathbf{H}^{(1)}_t[i] + \mathbf{H}^{(2)}_t[i],}&{{\rm{if  }}MR{I_t}[i] \ne 0}\\
{\mathbf{H}^{(1)}_t[i],}&{{\rm{if  }}MR{I_t}[i] = 0}
\end{array}} \right.
\label{eq_score}
\end{equation}

For each eviction decision, we first compute $\mathbf{I}_t$ with Eq.~\ref{eq_score}, then retain the most recent $W$ and $\rm{Top}(B-W)$ KV pairs with the highest scores.

Notably, LazyEviction strategy introduces additional computational overhead, which may affect inference efficiency. We analyze this in Appendix~\ref{cost} through both theoretical analysis and empirical evaluation. We point out that LazyEviction mitigates additional costs by making eviction decisions at fixed intervals rather than continuously. This $W$-steps eviction mechanism results in significantly lower overhead compared to existing KV compression methods. Moreover, since LazyEviction limits the KV budget during decoding steps, as the number of generated tokens increases, the computational efficiency does not grow linearly like it does with Full KV. When the generated tokens reach 16k, the overall inference efficiency of LazyEviction surpasses Full KV.



\paragraph{Disscussion about the choice of $W$ and $\alpha$.}

In LazyEviction, the observation window with size $W$ serves to detect recurring tokens. However, since the most recent $W$-th tokens must be preserved, the setting of $W$ inherently introduces a trade-off between local and global information retention. An excessively large $W$ risks losing critical past tokens, while an overly small $W$ may fail to detect numerous recurring tokens. To resolve this dilemma, we employ the observation from Fig.~\ref{attn}(c), setting $W$ as the MRI threshold corresponding to the 80\% tokens. This threshold can be statistically determined through offline analysis by randomly selecting 1\% samples from the test datasets. Our experiments in Appendix~\ref{wsize} analyze how $W$ affects reasoning performance.

Since the selection of $\alpha$ influences the temporal distribution of token timestamps, both extremely large and small $\alpha$ hinder effective identification of tokens' importance variation. As the methodology for determining $\alpha$ has been thoroughly discussed in RaaS~\citep{hu2025efficient}, we directly adopt their approach in LazyEviction. We analyze the effect of setting different $\alpha$ in Appendix~\ref{alpha}.


\section{Experimental Results}

In this section, we adopt variants of the DeepSeek-R1 distilled model~\citep{guo2025deepseek}: DeepSeek-R1-Distill-Llama-8B (DS-Llama-8B) and DeepSeek-R1-Distill-Qwen-7B (DS-Qwen-7B), and Qwen-series reasoning model: Qwen3-4B~\citep{yang2025qwen3} and QwQ-32B~\citep{team2024qwen2}. The evaluations are carried on five reasoning benchmarks from three different domains, including mathematical reasoning (GSM8K~\citep{cobbe2021training}, MATH-500~\citep{hendrycks2021measuring}, AIME~\citep{aime}), Science QA (GPQA Diamond ~\citep{rein2024gpqa}) and programming (LiveCodeBench~\citep{jain2024livecodebench}). 

We compare LazyEviction with FullKV and four representative KV cache compression methods, including TOVA~\citep{oren2024transformers}, H2O~\citep{zhang2023h2o}, RaaS~\citep{hu2025efficient}, and R-KV~\cite{cai2025r}. We use the KV compression ratio $r$ to denote the KV cache budget across different baselines, where $r$ represents the proportion of KV cache usage relative to FullKV. More implementation details can be found in Appendix~\ref{setup}.


\subsection{Performance Comparison with Baselines}

Table~\ref{main_res} and Table~\ref{res2} present a comprehensive performance comparison of our LazyEviction and baselines. The results demonstrate that LazyEviction consistently outperforms other methods across all datasets, achieving performance comparable to the baseline FullKV with minimal deviation. In mathematical reasoning tasks, LazyEviction achieves performance close to FullKV with only 30\%\textasciitilde50\% KV budgets. Additionally, LazyEviction also achieves the best performance across GPQA Diamond and LiveCodeBench datasets, demonstrating its applicability among different domain tasks. It is worth noting that even though R-KV achieves performance close to LazyEviction in mathematical reasoning tasks, its performance significantly declines on other datasets. This is due to R-KV's excessive reliance on the assumption that there are many similar tokens present in the reasoning path. In other tasks, similar tokens are not as prevalent as in mathematical tasks.


\begin{table}[t]
  \centering
  \small
  \caption{Performance of our LazyEviction compared with baselines on mathematical reasoning dataset with DS-Llama-8B, DS-Qwen-7B, Qwen3-4B and QwQ-32B. The best results among all methods are in \textbf{bolded}.}
    \begin{tabular}{l|c|c|c|c}
    \toprule
    Methods & DS-Llama & DS-Qwen & Qwen3 & QwQ \\
    \midrule
    \multicolumn{5}{c}{\textit{GSM8K (Compression Ratio $r=50\%$)}} \\
    \midrule
    FullKV & 81.73 & 89.92 & 93.32 & 95.61 \\
    \midrule
    RaaS  & 78.01  & 85.37 & 90.22 & 84.31 \\
    H2O    & 77.16 & 87.04 & 88.47 & 89.23 \\
    TOVA  & 72.25 & 80.52 & 85.13 & 72.10 \\
    R-KV  & 78.69 & 88.33 & \textbf{91.65} & \textbf{93.63}\\
    \rowcolor[rgb]{ .751,  .751,  .751} \textbf{Ours} & \textbf{80.06} & \textbf{88.40}  & 91.50  & 93.48 \\
    \midrule
    \multicolumn{5}{c}{\textit{MATH-500 (Compression Ratio $r=50\%$)}} \\
    \midrule
    FullKV & 74.8  & 86.0    & 87.2  & 87.2 \\
    \midrule
    RaaS  & 71.2  & 82.4  & 84.0    & 80.4 \\
    H2O   & 67.2  & 80.4  & 79.6  & 80.4 \\
    TOVA  & 69.6  & 74.8 & 77.6  & 70.4 \\
    R-KV  & 73.2 & 83.6  & 83.4 & 85.2 \\
    \rowcolor[rgb]{ .751,  .751,  .751} \textbf{Ours} & \textbf{75.2}  & \textbf{85.4}   & \textbf{85.8}  & \textbf{85.4} \\
    \midrule
    \multicolumn{5}{c}{\textit{AIME (Compression Ratio $r=30\%$)}} \\
    \midrule
    FullKV & 30.0    & 46.7  & 60.0    & 73.3 \\
    \midrule
    RaaS  & 23.3  & 36.7  & 46.7  & 53.3 \\
    H2O    & 26.7  & 33.3  & 40.0    & 53.3 \\
    TOVA  & 23.3  & 36.7  & 40.0    & 36.7 \\
    R-KV  & 26.7 & \textbf{43.3} & 46.7 & 56.7\\
   \rowcolor[rgb]{ .751,  .751,  .751} \textbf{Ours} & \textbf{30.0}    & \textbf{43.3}  & \textbf{53.3}  & \textbf{66.7} \\
    \bottomrule
    \end{tabular}%
  \label{main_res}%
\end{table}%


\begin{table}[t]
  \centering
  \small
  \caption{Performance of our LazyEviction compared with baselines on GPQA Diamond and LiveCodeBench with DS-Llama-8B and DS-Qwen-7B. The best results among all methods are in \textbf{bolded}.}
    \begin{tabular}{l|c|c}
    \toprule
    Methods & DS-Llama-8B & DS-Qwen-8B \\
    \midrule
    \multicolumn{3}{c}{\textit{GPQA Diamond ($r=50\%$)}} \\
    \midrule
    FullKV & 37.4  & 55.7 \\
    \midrule
    RaaS  & 29.9  & 45.4  \\
    H2O   & 26.8  & 42.1  \\
    TOVA  & 30.9  & 44.3  \\
    R-KV & 28.2 & 41.2  \\
    \rowcolor[rgb]{ .751,  .751,  .751} \textbf{LazyEviction} & \textbf{36.9}  & \textbf{54.6}  \\
    \midrule
    \multicolumn{3}{c}{\textit{LiveCodeBench ($r=40\%$)}} \\
    \midrule
    FullKV & 58.62 & 55.17 \\
    \midrule
    RaaS  &    53.45   & 39.66  \\
    H2O   &    53.45   &  50.00 \\
    TOVA  &    51.72   &  29.31 \\
    R-KV & 48.28 &  46.55 \\
    \rowcolor[rgb]{ .751,  .751,  .751} \textbf{LazyEviction} &  \textbf{56.90}   & \textbf{51.72} \\
    \bottomrule
    \end{tabular}%
  \label{res2}%
\end{table}%

\begin{figure*}[t]
  \centering
    \centering{\includegraphics[scale=0.27]{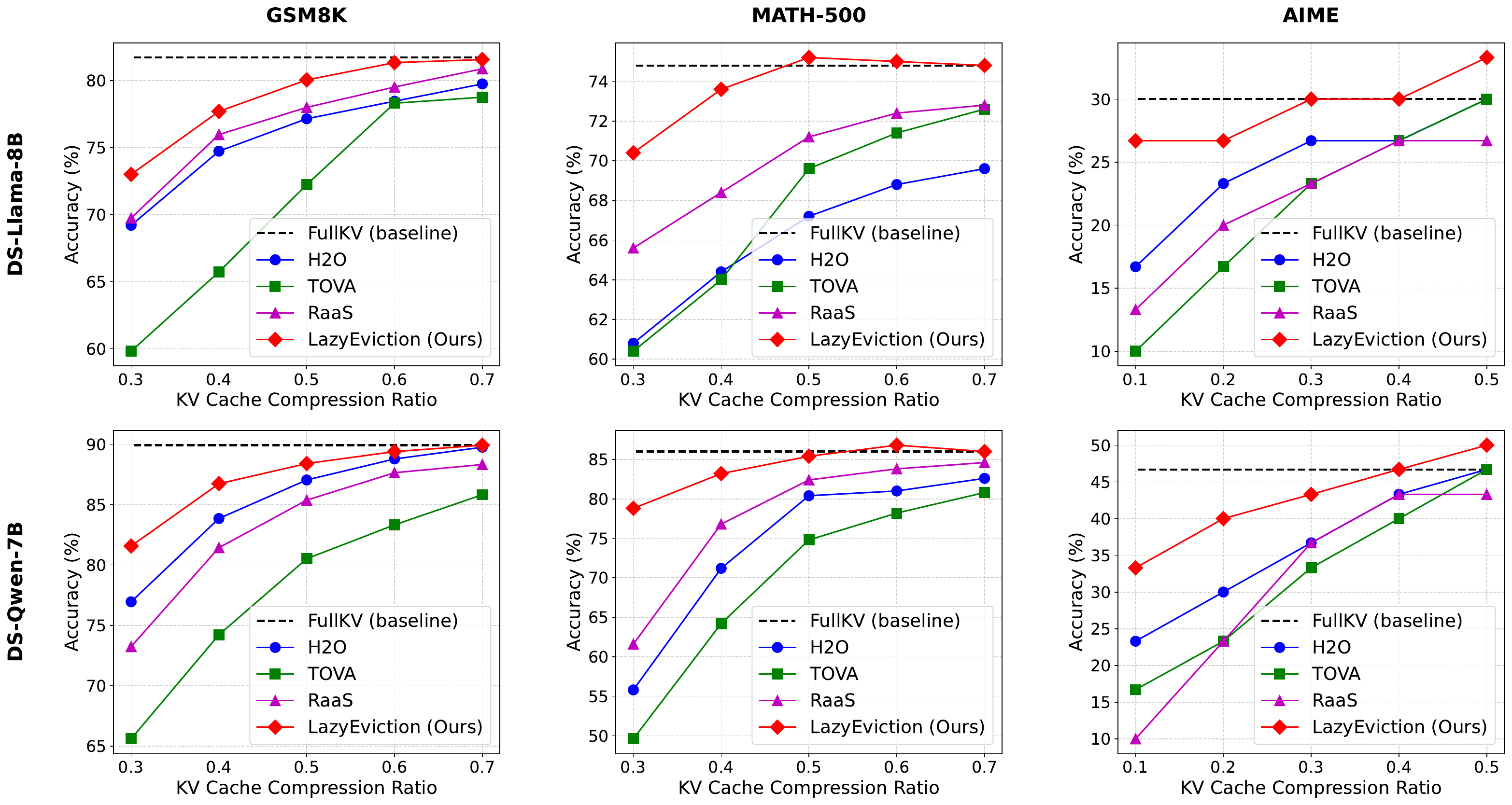}}
    \caption{Trade-off between accuracy and KV Cache among different datasets and models.}
 \label{memory}
\end{figure*}

\paragraph{Trade-off between Accuracy and KV Cache.}
We evaluate the performance of LazyEviction in optimizing the trade-off between KV cache budget and accuracy compared to baselines. As shown in Fig.~\ref{memory}, our method maintains higher accuracy across various budgets for both datasets and models. When the cache budget is large, other methods perform slightly worse than our approach. However, with a smaller cache budget, other methods experience substantial accuracy degradation. By implementing lagged KV eviction and leveraging MRI to track recurring tokens, our approach consistently achieves optimal performance, even under stringent budget constraints. Notably, on MATH-500 and AIME datasets, LazyEviction even outperformed FullKV's performance while reducing the KV cache budget by 50\%.

\subsection{Memory Efficiency of LazyEviction}


\begin{figure}[t]
    \vspace{-0.5cm}
    \centering
    \includegraphics[width=2.5in]{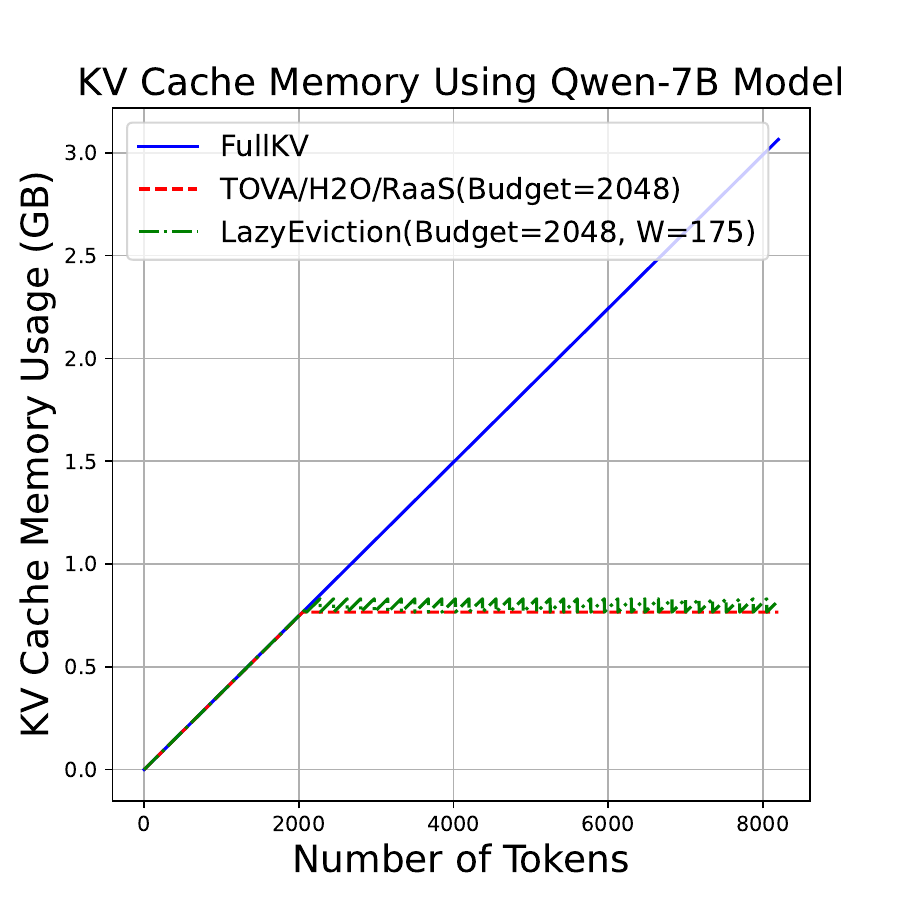}
	\caption{KV cache memory usage of different algorithms with varying output length (0-8k tokens).}
    \label{kv_var}
\end{figure}

We evaluate the KV cache memory usage of different algorithms on DS-Qwen-7B model, analyzing its memory usage variation with output token length. As shown in Fig.~\ref{kv_var}, the memory consumption of FullKV increases linearly with the number of tokens. In contrast, the memory usage of TOVA, H2O, and RaaS algorithms remains constant once the token count exceeds the budget. LazyEviction, with the observation window mechanism, exhibits minor fluctuations in KV cache memory usage after surpassing the budget, but Fig.~\ref{kv_var} indicates that its memory growth is minimal compared to the other three algorithms.

\subsection{Ablation Study}
\begin{table}[t]
  \centering
  \small
  \caption{Baseline algorithms with observation window mechanism, evaluating on the GSM8K dataset using DS-Llama-8B with $r=50\%$ and $W=25$.}
    \begin{tabular}{r|cr}
    \toprule
    \multicolumn{1}{l|}{Model} & DS-Distill-Llama-8B &  \\
    \midrule
    \multicolumn{1}{l|}{LazyEviction} & 80.06 &  \\
    \midrule
    \multicolumn{1}{l|}{H2O} & 77.16 &  \\
    + window &    78.88 (+1.72)   &  \\
    \midrule
    \multicolumn{1}{l|}{TOVA} & 72.25 &  \\
    + window &    76.09 (+3.84)   &  \\
    \midrule
    \multicolumn{1}{l|}{RaaS} & 78.01 &  \\
    + window &    78.92 (+0.91)   &  \\
    \bottomrule
\end{tabular}%
  \label{window}%
\end{table}

\paragraph{Lagged Eviction with an Observation Window}
Since our method retains a slightly increased number of KV pairs within the observation window, which may lead to the perceived performance gains being predominantly attributed to these additional tokens. To eliminate the impact of the additional tokens within observation window, we explore the other algorithms with the observation window mechanism on performance. As shown in Table~\ref{window}, when H2O, TOVA, and RaaS algorithms perform KV eviction every $W$ decoding steps, their performance improves. This improvement arises from the linear increase in retained KV pairs between consecutive decision decoding steps. However, their performance remains inferior to that of LazyEviction. LazyEviction employs MRI to track recurring tokens, effectively capturing their potential future importance, which leads to superior performance.

\paragraph{The Discussion of the Importance Score}
\label{score}
 We further investigate the influence of MRI-Centric Importance Score, as shown in Table~\ref{score_table}. If the importance score $\mathbf{I}_t$ calculation excludes the \textbf{H1-Score}, accuracy significantly decreases, underscoring the effectiveness of the recurrence interval tracking mechanism. Similarly, omitting the \textbf{H2-Score} leads to a slight accuracy drop, indicating that tokens with smaller MRI values are indeed more likely to become important in the future. Collectively, these findings demonstrate the effectiveness of LazyEviction’s MRI-Centric Eviction Policy.

  \begin{table}[t]
 \small
  \centering
  \caption{Performance with different importance scores.}
    \begin{tabular}{r|c|c}
    \toprule
    Model & DS-Llama-8B & DS-Qwen-7B \\
    \midrule
    \multicolumn{1}{l|}{LazyEviction} & 80.06 & 88.40\\
    \midrule
    w/o the \textbf{H1-Score} &  76.11 (-3.95) & 82.78 (-5.62)\\
    w/o the \textbf{H2-Score} & 79.67 (-0.39) & 87.21 (-1.19)\\
    \bottomrule
    \end{tabular}%
  \label{score_table}%
\end{table}

\section{Conclusion}

In this work, we address the critical bottleneck of KV cache memory overhead in long reasoning tasks. Our analysis reveals the Token Importance Recurrence phenomenon, where certain significant tokens receive renewed attention after multiple decoding steps. Based on this insight, we propose LazyEviction, a novel lagged eviction framework that introduces temporal observation windows, recurrence interval tracking, and an MRI-centric eviction policy to preserve recurring tokens while dynamically reducing memory usage. Experimental results show that LazyEviction achieves performance close to or outperforms FullKV with only 30\%\textasciitilde50\% KV budgets, highlighting the importance of preserving recurring tokens on maintaining knowledge continuity in reasoning tasks. By shifting from greedy eviction to predictive retention, our work provides a new paradigm for efficient KV cache compression in long reasoning scenarios.

\section*{Limitations}

\paragraph{Dynamically Adjusting in Observation Window Size.} We discuss in Appendix~\ref{wsize} that the suitable observation window size is crucial for identifying recurring tokens. In Sec.~\ref{observation}, we find that the MRI distribution of recurring tokens varies across different models and tasks. Therefore, we propose to offline select 1\% of samples to pre-statistically analyze the MRI distribution to determine the observation window size. However, this approach is inefficient and sub-optimal in practice. Thus, dynamically and adaptively adjusting the observation window based on task characteristics will be an important direction for improvement in LazyEviction in the future.

\paragraph{Evaluation Beyond Reasoning Tasks.} Although we have conducted experiments on tasks from different domains, our evaluation is still limited to reasoning tasks. This is because the core finding of this paper, the Token Importance Recurrence, is particularly significant in reasoning scenarios. We attempt to investigate TIR in non-reasoning scenarios. For example, we use Llama-3.1-8B-Instruct\footnote{https:https://huggingface.co/meta-llama/Llama-3.1-8B-Instruct} to analyze the MRI distribution on the language modeling dataset C4\footnote{https://huggingface.co/datasets/allenai/c4}. The results show that although TIR also exists, recurring tokens have a relatively smaller MRI (<10). In this case, the performance of LazyEviction is not significantly different from that of Cumulative Attention-based Eviction methods (e.g., H2O, RaaS). Therefore, we clarify that the observations and methods presented in this paper are more applicable to reasoning scenarios.

\paragraph{Evaluation on Larger Scale Models.} Our experiments cover reasoning models ranging from 4B to 32B in size. However, due to limitations in the experimental environment, we have not explored larger scale models, such as Qwen3-Max and DeepSeek-R1. Inference on models at the 100B level is extremely time- and resource-consuming. For example, evaluating 500 samples MATH-500 with DeepSeek-R1 requires days on 8 A100 GPUs. Nonetheless, we believe that our core finding, the Token Importance Recurrence phenomenon, and the importance of preserving recurring tokens exist across all reasoning models, and we will extend our experiments to more models in the future.

\bibliography{custom}

\appendix

\section{Mathematical Formulation of LazyEviction}
\label{formulation}

Suppose the sequence length is $N_t$ at decoding step $t$. We denote $\mathbf{Q}_i$, $\mathbf{K}_i$ and $\mathbf{V}_i$ as the query, key and value vectors of the $i$-th token. The attention output at the decoding step $t$ can be formulated as:

\begin{equation}
\mathbf{A}_t({\mathbf{Q}_{{N_t}}},\mathbf{K},\mathbf{V}) = {\rm{Softmax}}\left( {\frac{{\mathbf{Q}_{N_t}{\mathbf{K}^T}}}{{\sqrt {{d_h}} }}} \right)\mathbf{V},
\label{eq_attn}
\end{equation}

where $\mathbf{K} = [\mathbf{K}_1, ..., \mathbf{K}_{N_t}] \in \mathbb{R}^{N_t \times d_h}$, $\mathbf{V} = [\mathbf{V}_1, ..., \mathbf{V}_{N_t}] \in \mathbb{R}^{N_t \times d_h}$ and $d_h$ is the head dimension.  We define $\mathcal{S}_t$ as the set of the KV pairs $(\mathbf{K}_{i}, \mathbf{V}_{i})$ for all the previous tokens $i<N_t$ cached in GPU memory at decoding step $t$. Since the size of $\mathbf{K}$ and $\mathbf{V}$ grows linealy with the decoding step $t$, current methods try to evict the unimportant KV pairs from $\mathcal{S}_t$ at each step $t$. Unlike the previous KV compression methods, we make KV eviction decisions at decoding step $t = kW$ ($k \in \mathbf{N}^+$). Denote $\mathcal{S}^{*}_t$ as the optimal KV pairs selected from $\mathcal{S}_t$ under the cache budget $B$.  Thus, the KV eviction problem can be formulate as:

\begin{multline}
\mathcal{S}_t^*=\mathop {\arg \min }\limits_{\left| {\mathcal{S}_t'} \right| = B, {\mathcal{S}_t'} \subseteq {\mathcal{S}_t}} \\
{\left\| {\mathbf{A}_t}({\mathbf{Q}_{{N_t}}},\mathbf{K},\mathbf{V}) - {\mathbf{A}_t'}({\mathbf{Q}_{{N_t}}},\mathbf{K'},\mathbf{V'},{\mathcal{S}'_t}) \right\|}_2
\label{eq_opt}
\end{multline}

where $\mathbf{K'}$ and $\mathbf{V'}$ are selected KV pairs given by $\mathcal{S}'_t$, $\mathbf{A}_t'$ is the attention output computed by Eq.~\ref{eq_attn} after KV eviction. To minimize the effect of the KV eviction process, our goal is to find the optimal $\mathcal{S}^{*}_t$ to minimize the change in attention output. 

It is intractable to directly solve Eq.~\ref{eq_opt} because of its combinatorial nature. We design a heuristic method to efficiently find the suboptimal solution for $\mathcal{S}^{*}_t$. First, as emphasized in previous works~\cite{xiao2023efficient, zhang2023h2o, ghadia2025dialogue}, retaining recent tokens is critical to ensuring coherence in the generated context. And to observe the importance variations of the newly generated tokens, we consistently preserve the latest $W$ KV pairs. Second, among the remaining past KVs, we retain $B-W$ KV pairs by prioritizing tokens most likely to exhibit importance recurrence in future steps. This selection is guided by an auxiliary score vector $\mathbf{I}_t$, which quantifies the potential future importance of tokens. Formally, our eviction policy is defined as:

\begin{equation}
\mathcal{S}_t' = \rm{Top}_{[B-W]}(\mathbf{I}_t) \cup \mathcal{W}_t
\label{eq_evic}
\end{equation}

where $\rm{Top}_{[k]}(\cdot)$ represents the selection of top-k entries (KV pairs) from $\mathcal{S}_t$ and $\mathcal{W}_t$ is the set of the most $W$ recent KV pairs in $\mathcal{S}_t$.

\vspace{-0.5cm}
\begin{center}
\begin{minipage}{8cm}
 \begin{algorithm}[H]
    \caption{LazyEviction}
    \begin{algorithmic}[1]
        \label{code}
        \REQUIRE $\mathbf{A}_t$, $\mathbf{K}$, $\mathbf{V}$, $W$, $B$, $\alpha$, $t$.
        \FOR { All token $i$}
            \IF {$A_t[i]>=\alpha$}
                \STATE Update $TS_t[i] = t$.
            \ENDIF
        \ENDFOR
        \STATE Update $\mathbf{MRI}_t$ according to Eq.~\ref{eq_MRI}.
        \STATE Calculate $\mathbf{I}_t$ according to Eq.~\ref{eq_score}.
            
        \IF {$t == kW$ and $B>=\left| {\mathbf{K}} \right|$}
            \STATE Update KV pairs according to Eq.~\ref{eq_evic}.
        \ENDIF
        \RETURN $\mathcal{S}_t'$.
    \end{algorithmic}
\end{algorithm}
\end{minipage}
\end{center}

\section{Overall Procedure of LazyEviction}
\label{algo}

We provide the pseudo-code of LazyEviction in Algorithm.~\ref{code}, which contains \textbf{Recurrence Interval Tracking} at each step $t$ and \textbf{MRI-Centric Eviction} at the decoding step $t = kW$ ($k \in \mathbf{N}^+$). Thus, Lazyeviction tasks the current decoding step $t$, observation window size $W$, cache budget $B$, $\alpha$, and $\mathbf{A}_t$, $\mathbf{K}$, $\mathbf{V}$ as input.

\section{Experiment Setup}
\label{setup}

\paragraph{Models and Datasets} By fine-tuning with CoT data generated by DeepSeek-R1, even smaller distilled models can acquire long-reasoning capabilities similar to DeepSeek-R1~\citep{guo2025deepseek}. Thus, we first adopt DeepSeek-R1-Distill-Llama-8B and DeepSeek-R1-Distill-Qwen-7B as the base models for evaluating KV cache compression.  In addition, to validate the effectiveness of LazyEviction across a broader range of model types, we also utilize additional Qwen-series reasoning models with different sizes: Qwen3-4B~\citep{yang2025qwen3} and QwQ-32B~\citep{team2024qwen2}.


The evaluations are mainly carried on three widely used mathematical reasoning benchmarks: GSM8K\citep{cobbe2021training}, MATH-500\citep{hendrycks2021measuring} and AIME~\citep{aime}. GSM8K is a high-quality, linguistically diverse dataset of elementary-level math problems requiring multi-step reasoning and basic arithmetic operations to solve. MATH-500 contains 500 challenging problems drawn from high school math competitions, spanning domains such as algebra, geometry, and number theory. AIME is a math problem dataset collected from the American Invitational Mathematics Examination, designed to challenge the most exceptional high school mathematics students in the United States. 

Besides mathematical reasoning benchmarks, we also adopted reasoning benchmarks from other domains to validate the generalization of LazyEviction across different domain tasks. Specifically, GPQA Diamond ~\citep{rein2024gpqa} is a multiple-choice science QA dataset covering the fields of biology, physics and chemistry, while LiveCodeBench~\citep{jain2024livecodebench} is an advanced evaluation benchmark (1116 samples) used to rigorously assess the code processing capabilities of LLMs. Since it may cost days to evaluate the entire benchmark, we randomly sample 5\% data (58 samples) from the dataset for evaluation.


\paragraph{Baselines} To validate the performance of LazyEviction, we compare it against Full Cache and three representative KV cache compression methods, including TOVA~\citep{oren2024transformers}, which retains tokens with the highest attention scores at each decoding step; H2O~\citep{zhang2023h2o}, which preserves heavy hitter tokens (identified by cumulative attention scores) and recent tokens during decoding; RaaS~\citep{hu2025efficient}, which selects tokens with the newest timestamps at every decoding step; and R-KV~\citep{cai2025r}, which utilize the similarity among tokens to identify and evict redundant tokens in long reasoning tasks.

\paragraph{Implementation Details} We implement LazyEviction in HuggingFace transformers library ~\citep{wolf2020transformers} to integrate it into the existing attention mechanism. We set the observation window size equal to the value of 80\% MRI threshold based on the preliminary observations (Fig.~\ref{attn}(c)), and the number of recent tokens in H2O is equal to LazyEviction's window size. Unless stated otherwise, we set $\alpha = 0.0005$ for DS-Llama-8B model and $\alpha = 0.0001$ for DS-Qwen-7B, Qwen-4B and QwQ-32B models. Experiments are conducted on NVIDIA V100 (32GB) GPUs. We use Ubuntu 20.04 with Linux kernel 6.8.0 and CUDA 12.8. 
The maximum number of generated tokens are set as 4096 for GSM8K dataset, 8192 for MATH-500 and GPQA datasets, 16384 for AIME and LiveCodeBench datasets, respectively.




\section{Discussion about formulations of the importance score.}
\label{Appen_formulation}

As illustrated in Sec~\ref{sec_lazy}, both H1-score and H2-score are formulated by sigmoid functions. The choice of the formulations have been thoroughly considered, where the score function must possess the following two properties: (1) The score function needs to have a monotonically decreasing property; (2) The score values must fall within the range of [0, 1]. For functions that meet these criteria, we have tried exponential, logarithmic, inverse, sigmoid, and tanh functions. In preliminary experiments, we found that the sigmoid function yielded the most stable results and was more concise in form, leading us to define the importance score in Sec~\ref{sec_lazy}.

We evaluated different combinations of H1 and H2 score functions on the GSM8K and MATH-500 datasets. As shown in Table~\ref{table_formulation}, each function forms perform very similarly. Among them, the sigmoid functions achieves a better and stable performance. Therefore, we adopt sigmoid function for H1-score and H2-score.

\begin{table}[h]
  \centering
  \small
  \caption{Measurement of average decoding latency and throughput for LazyEviction compared with baselines.}
    \begin{tabular}{l|l|l}
    \toprule
     & H1-score & H2-score  \\
    \midrule
    \multicolumn{3}{c}{\textit{GSM8K}} \\
    \midrule
    Initial	&  Sigmoid: 88.40 	& Sigmoid: 88.40 \\
    \midrule
    Change to	&  Exp: 88.25 (-0.15)	&   Exp: 88.25 (-0.15) \\
        & Tanh: 88.25 (-0.15)	& Tanh: 88.33 (-0.07) \\
        & Log: 88.40 (-)	& Log: 88.33 (-0.07) \\
        & Inverse: 88.40 (-)	& Inverse: 88.40 (-) \\
    \midrule
    \multicolumn{3}{c}{\textit{MATH-500}} \\
    \midrule
    Initial	& Sigmoid: 85.4 & Sigmoid: 85.4  \\
    \midrule
    Change to	& Exp: 85.2	(-0.2)	& Exp: 85.4 (-) \\
        & Tanh: 85.2 ((-0.2)	& Tanh: 84.2 (-0.8) \\
        & Log: 85.0 (-0.4)	& Log: 84.8 (-0.6) \\
        & Inverse: 85.0 (-0.4)	&  Inverse: 85.2  (-0.2)\\
    \bottomrule
    \end{tabular}%
  \label{table_formulation}%
\end{table}%

\section{Computational Cost of LazyEviction}
\label{cost}
It is important to notice that LazyEviction introduces additional computational overhead, which may affect inference efficiency. Therefore, it is necessary to analyze the additional computational introduced by LazyEviction. We first clarify that all KV eviction strategies inherently incur computational overhead, as they require decision-making at each decoding step. However, LazyEviction mitigates this cost by performing eviction decisions at fixed intervals, rather than continuously. This $W$-steps eviction mechanism leads to significantly lower overhead compared to baseline methods, which we will demonstrate through both theoretical analysis and empirical evaluation.

\subsection{Theoretical Analysis of Computational Complexity}

Unlike existing works that perform KV eviction decisions at every decoding step, LazyEviction’s Observation Window-based Lagged Eviction adopts a window $W$ as the interval for KV eviction decisions. Therefore, we analyze computational complexity over $W$ decoding steps. All KV cache eviction algorithms can be divided into two steps: importance score calculation and importance ranking. We only consider the additional computational overhead, where the computation of attention process is not involved. Given a KV cache budget B, the comparison between LazyEviction and baseline methods is shown in Table~\ref{complexity}.

 \begin{table}[h]
 \small
  \centering
  \caption{Computational complexity for each eviction algorithms within one window.}
    \begin{tabular}{l|l}
    \toprule
    Method &	Computational Complexity \\
    \midrule
    H2O	& $O[W(B+BlogB)]$ \\
    TOVA	& $O(WBlogB)$ \\
    RaaS	& $O[W(B+BlogB)]$ \\
    LazyEviction	& $O(WB+BlogB)$ \\
    \bottomrule
    \end{tabular}%
  \label{complexity}%
\end{table}

Here, $BlogB$ is the computational complexity of a single importance ranking. Compared to LazyEviction, baseline methods require W importance rankings within $W$ steps. For importance score calculation, LazyEviction, like other baseline methods, requires a finite number of computations for $B$ KV pairs (Eq~\ref{eq_MRI} MRI updating and Eq~\ref{eq_score} Importance score calculating). Since $W\gg1$, the computational overhead of LazyEviction is minimal compared to baseline methods.

\subsection{Experimental Computational Overhead Measurement}

\paragraph{Single-Step Decoding Latency}

For FullKV, single-step decoding latency grows linearly with the generation length (KV cache increases linearly). In contrast, KV cache eviction can maintain decoding latency at a constant level. The following table shows the comparison of single-step decoding latency measured at different positions during the generation of a sequence with a length of 16k (KV cache budget set to 8192, $r=50\%$). 
As shown in Table~\ref{latency}, the results demonstrate that although LazyEviction introduces additional computational overhead (latency is greater when output length is less than budget), it effectively reduces single-step decoding latency by shortening the KV sequence length (reduce to 8k), which is very important for long reasoning scenarios.

\begin{table}[h]
  \centering
  \small
  \caption{Measurement of single-step decoding latency (ms) for LazyEviction compared with FullKV.}
    \begin{tabular}{l|ccccc}
    \toprule
     Step  & 2K & 4K   & 8K  & 12K & 16K \\
    \midrule
    FullKV	& 16.46	& 16.91	 & 18.01 & 20.76	& 27.14 \\
    \midrule
    LazyEviction	& 17.81 & 17.89	& 21.56	& 22.03	& 22.52 \\
    \bottomrule
    \end{tabular}%
  \label{latency}%
\end{table}%

\paragraph{Average Decoding Latency and Throughput}

We illustrate that LazyEviction introduces the least additional computational overhead by comparing average decoding latency and throughput. Since TOVA does not introduce any additional computations aside from importance scoring, it has the smallest computational overhead among methods other than LazyEviction, so we mainly compared it with the TOVA method. As shown in Table~\ref{throughtput}, LazyEviction outperforms TOVA at different output lengths, indicating that our LazyEviction introduces less additional computational overhead. Moreover, LazyEviction outperforms when Generation Length=16K, indicating LazyEviction's computational efficiency.

\begin{table}[h]
  \centering
  \small
  \caption{Measurement of average decoding latency and throughput for LazyEviction compared with baselines.}
    \begin{tabular}{l|c|c|c}
    \toprule
    Methods & Budget & \makecell[c]{Throughput \\ (token/s) $\uparrow$ } & \makecell[c]{Avg. Latency \\(ms/token) $\downarrow$}  \\
    \midrule
    \multicolumn{4}{c}{\textit{Generation Length=4K}} \\
    \midrule
    FullKV	& -	& 60.19	& 16.61 \\
    TOVA	& 2048 	& 46.11	& 21.69 \\
    LazyEviction	& 2048 	& 55.94	& 18.99 \\
    \midrule
    \multicolumn{4}{c}{\textit{Generation Length=8K}} \\
    \midrule
    FullKV	& - &	57.90	& 17.27 \\
    TOVA	& 4096 	& 45.62	& 21.92 \\
    LazyEviction	& 4096 	& 51.75	& 19.32 \\
    \midrule
    \multicolumn{4}{c}{\textit{Generation Length=16K}} \\
    \midrule
    FullKV	& -	& 47.61	& 21.07 \\
    TOVA	& 8192 	& 43.94 & 	22.76 \\
    LazyEviction	& 8192 & 	48.64 & 	20.56 \\
    \bottomrule
    \end{tabular}%
  \label{throughtput}%
\end{table}%

\section{Discussion about $W$ and $\alpha$}

\subsection{The Selection of Window Size $W$.}
\label{wsize}
We investigate the impact of the window size $W$ on DS-Llama-8B performance across two datasets in Table~\ref{window_size}. As $W$ increases, the model's accuracy generally improves. This is because a larger window size allows the LazyEviction mechanism's MRI metric to detect more recurring tokens, resulting in better capturing a token's potential future significance, thereby enhancing performance. However, when the window size becomes too large, LazyEviction risks discarding critical past tokens, leading to a slight performance decline. Thus, the choice of $W$ presents a trade-off between retaining local and global information. 

\begin{table}[h]
  \centering
  \small
  \caption{Performance of different setting of the window size $W$ on DS-Llama-8B.}
    \begin{tabular}{c|ccccc}
    \toprule
    \textbf{Dataset} & \multicolumn{5}{c}{\textbf{GSM8K ($r=50\%$)}} \\
    \midrule
          & W=4 & W=8   & W=16  & \textbf{W=25} & W=32 \\
    \midrule
    Acc.  & 75.14 & 76.58 & 76.73 & \textbf{80.06} & 79.63 \\
    \midrule
    \textbf{Dataset} & \multicolumn{5}{c}{\textbf{MATH-500 ($r=50\%$)}} \\
    \midrule
           & W=8   & W=16  & W=32  & \textbf{W=52} & W=64 \\
    \midrule
    Acc. & 71.8  & 72.2  & 72.8  & \textbf{75.2} & 74.8 \\
    \bottomrule
    \end{tabular}%
  \label{window_size}%
\end{table}%


\subsection{The Selection of $\alpha$}
\label{alpha}
We further explore the impact of the threshold $\alpha$ on the GSM8K dataset using two models. The value of $\alpha$ influences the MRI of each token, with $\alpha=0.0005$ producing optimal results for DS-Llama-8B and $\alpha=0.0001$ yielding the best performance for DS-Qwen-7B, as detailed in Table \ref{tab:addlabel4}. First, when $\alpha$ is small, tokens are more readily flagged as important, but the MRI of these tokens is reduced, which does not accurately reflect the periodicity of their significance. Second, when $\alpha$ is large, the steps where tokens gain importance are not captured, preventing the detection of their recurring patterns.

\begin{table}[h]
  \centering
  \small
  \caption{Performance of different setting of $\alpha$.}
    \begin{tabular}{c|c|ccc}
    \toprule
    \textbf{Model} & \multicolumn{4}{c}{\textbf{DS-Llama-8B ($r=50\%$)}} \\
    \midrule
          & FullKV & $\alpha$=0.0001 & $\alpha$=0.0005 & $\alpha$=0.001 \\
    \midrule
    Acc. & 81.73 &    78.97 & \textbf{80.06} &  79.68 \\  
    \midrule
    \textbf{Model} & \multicolumn{4}{c}{\textbf{DS-Qwen-7B ($r=50\%$)}} \\
    \midrule
          & FullKV & $\alpha$=0.00001 & $\alpha$=0.0001 & $\alpha$=0.001 \\
    \midrule
    Acc. & 89.92  &    88.06   & \textbf{88.40} &  88.32 \\ 
    \bottomrule
    \end{tabular}%
  \label{tab:addlabel4}%
\end{table}%

\section{Checklist Issues}

The datasets used in the experiment, including GSM8k (MIT), MATH500 (MIT), AIME (MIT), GPQA Dimond (MIT), Livecodebench (MIT) and models, including  DeepSeek-R1-Distill-Llama-8B(MIT), DeepSeek-R1-Distill-Qwen-7B (MIT), Qwen3-4B (apache2.0), QwQ-32B (apache2.0), are used with their intended usage scenarios.
We retrieve all models and datasets from HuggingFace\footnote{https://huggingface.co/}, where detailed documentation, including parameter sizes and model architectures, is provided. We manually checked the data and believe there is no personal information misused.

We used ChatGPT for paper editing (e.g., grammar, spelling, word choice). 

To the best of our knowledge, we believe our work does not pose risks that harm any subgroup of our society.

\end{document}